\documentclass[journal]{IEEEtran}

\usepackage{times}
\usepackage{soul}
\usepackage{url}
\usepackage{hyperref}
\usepackage[utf8]{inputenc}
\usepackage[small]{caption}
\usepackage{graphicx}
\usepackage{amsmath}
\usepackage{amsthm}
\usepackage{booktabs}
\usepackage{algorithm}
\usepackage{algorithmic}
\usepackage{multirow}
\usepackage{subfigure}
\usepackage{stfloats}
\usepackage[switch]{lineno}

\begin{document}
%
\title{BSNN: Towards Faster and Better Conversion of Artificial Neural Networks to Spiking Neural Networks with Bistable Neurons}
%
%
%

\author{Yang Li, Yi Zeng, and Dongcheng Zhao
\thanks{Yang Li is with the Research Center for Brain-inspired Intelligence, Institute of Automation, Chinese Academy of Sciences (CASIA), Beijing 100190, China, and School of Artificial Intelligence, University of Chinese Academy of Sciences (UCAS), Beijing 100190, China, e-mail: liyang2019@ia.ac.cn}
\thanks{Dongcheng Zhao is with CASIA and UCAS, Beijing 100190, China.}
\thanks{Yi Zeng is with CASIA, UCAS and Center for Excellence in Brain Science and Intelligence Technology, Chinese Academy of Sciences, Shanghai 200031, China, e-mail: yi.zeng@ia.ac.cn}
\thanks{The corresponding author is Yi Zeng.}
\thanks{Manuscript received May 27, 2021; revised XX XX, 2021.}}

%
%

\markboth{LATEX}%
{Shell \MakeLowercase{\textit{et al.}}: Bare Demo of IEEEtran.cls for IEEE Journals}
%



\maketitle

\begin{abstract}
	The spiking neural network (SNN) computes and communicates information through discrete binary events. It is considered more biologically plausible and more energy-efficient than artificial neural networks (ANN) in emerging neuromorphic hardware. However, due to the discontinuous and non-differentiable characteristics, training SNN is a relatively challenging task. Recent work has achieved essential progress on an excellent performance by converting ANN to SNN. Due to the difference in information processing, the converted deep SNN usually suffers serious performance loss and large time delay. In this paper, we analyze the reasons for the performance loss and propose a novel bistable spiking neural network (BSNN) that addresses the problem of spikes of inactivated neurons (SIN) caused by the phase lead and phase lag. Also, when ResNet structure-based ANNs are converted, the information of output neurons is incomplete due to the rapid transmission of the shortcut path. We design synchronous neurons (SN) to help efficiently improve performance. Experimental results show that the proposed method only needs 1/4-1/10 of the time steps compared to previous work to achieve nearly lossless conversion. We demonstrate state-of-the-art ANN-SNN conversion for VGG16, ResNet20, and ResNet34 on challenging datasets including CIFAR-10 (95.16\% top-1), CIFAR-100 (78.12\% top-1), and ImageNet (72.64\% top-1).
\end{abstract}


%
\IEEEpeerreviewmaketitle

\section{Introduction}

Deep learning (or Deep Neural Network, DNN) has made breakthroughs in many fields such as computer vision \cite{redmon2016you,liu2016ssd,girshick2015fast}, natural language processing \cite{bahdanau2014neural,devlin2018bert}, and speech processing \cite{park2020improved}, and has even surpassed humans in some specific fields. But many difficulties and challenges also need to be overcome in the development process of deep learning \cite{kemker2018measuring,yan2019modeling,nguyen2015deep,lake2015human}. One concerning issue is that researchers pay more attention to higher computing power and better performance while ignoring the cost of energy consumption \cite{strubell2019energy}. Taking natural language processing tasks as an example, the power consumption and carbon emissions of Transformer \cite{vaswani2017attention} model training are very considerable. In recent years, the cost advantages and environmental advantages of low-energy AI have attracted the attention of researchers. They design compression algorithms \cite{he2018learning,wu2016quantized}to enable artificial neural networks (ANN) to significantly reduce network parameters and calculations while maintaining their original performance. Another part of the work focuses on computing architecture \cite{chen2014diannao}, less computational energy consumption can be achieved by designing hardware that is more suitable for the operational characteristics of neural network models. But the problem of the high computational complexity of deep neural networks still exists. Therefore, the spiking neural network, known as the third-generation artificial neural network \cite{maass1997networks}, has received more and more attention \cite{tavanaei2019deep,wang2020supervised,illing2019biologically,jang2019introduction,bing2018survey}.

Spike neural networks (SNNs) process discrete spike signals through the dynamic characteristics of spiking neurons, rather than real values, and are considered to be more biologically plausible and more energy-efficient \cite{roy2019towards,lobo2020spiking,pfeiffer2018deep}. For the former, the event-type information transmitted by neurons in SNN is the spike, which is generated when the membrane potential reaches the neuron firing threshold. Thus, its information processing process is more in line with biological reality than traditional artificial neurons \cite{zhang2018plasticity,liang2021stylistic,fang2021brain}. For the latter, the information in SNN is based on the event, e.g., neurons that do not emit spikes do not participate in calculations, and the information integration of neurons is an accumulate (AC) operation, which is more energy-efficient than the multiply-accumulate (MAC) operations in ANN \cite{zhao2014event,marian2002efficient}. Therefore, researchers put forward the concept of neuromorphic computing \cite{burr2017neuromorphic,davies2019benchmarks,song2020compiling}, which realizes the more biologically plausible SNN on hardware. It shows more significant progress in fast information processing and energy saving. But due to the non-differentiable characteristics of SNN, training SNN is still a challenging task. Because of the lack of the derivative of the output, the common backpropagation algorithm cannot be used directly. How to use SNN for effective reference has become a problem for researchers.

Taking inspiration from the brain, such as Spike-Timing Dependent Plasticity (STDP) \cite{bi1998synaptic,bengio2015stdp}, lateral inhibition \cite{abbott2000synaptic,blakemore1970lateral}, Long-Term Potentiation (LTP) \cite{malenka2003long}, and Long-Term Depression (LTD) \cite{ito1989long} are effective methods. By properly integrating different neural mechanisms in the brain \cite{zeng2017improving}, SNN can be effectively trained. Because most of these methods are unsupervised, researchers often add SVM \cite{noble2006support} or other classifiers for supervised learning \cite{wang2020supervised,hao2020biologically} or directly do learning in an unsupervised manner \cite{illing2019biologically,diehl2015unsupervised}. All of them are of great importance for further enhancing the interpretability of SNN and exploring the working mechanism of the human brain. However,  this optimization method that only uses local neural activities is challenging to achieve high performance and be applied to complex tasks. Some researchers try to train SNNs through approximated gradient algorithms \cite{wu2019direct,shrestha2018slayer,lee2020enabling,zhang2020temporal}, where the backpropagation algorithm can be applied to the SNN by continuous the spike firing process of the neuron. However, this method suffers from difficulty in convergence and requires a lot of time in training procedure in the deep neural networks (DNN) because it is difficult to balance the whole firing rate. For the above two methods, they perform poorly in large networks and complex tasks. We believe that the inability to obtain an SNN with effective reference ability is a key issue in the development and application of SNN.

\begin{figure}[t]
	\centering 
	\includegraphics[width=0.45\textwidth]{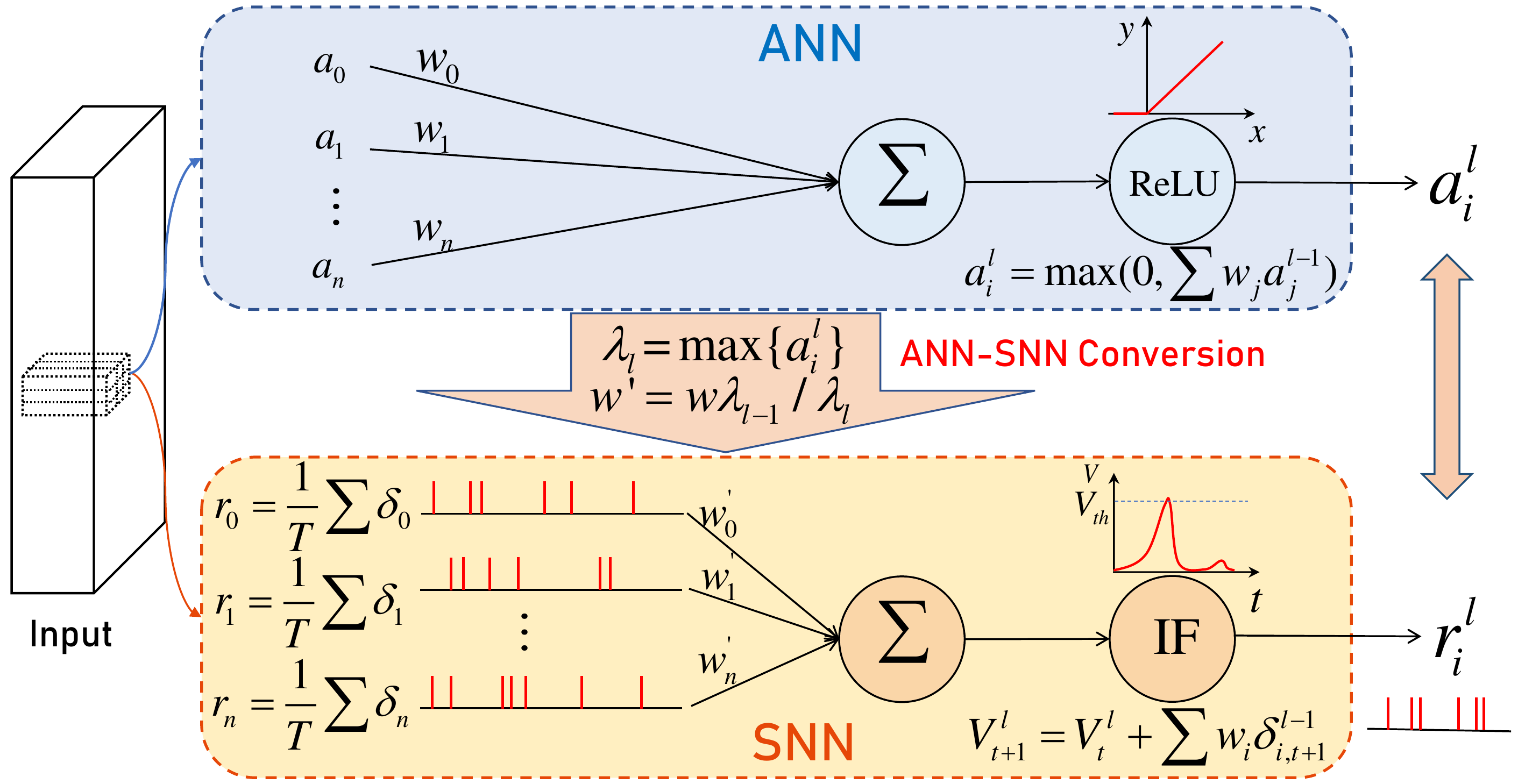}
	\caption{Illustration of ANN-SNN Conversion}
	\label{Fig.1}
\end{figure}

Recently, the conversion method has been proposed to convert the training result of ANN to SNN \cite{cao2015spiking}. The ANN-SNN conversion method maps the trained ANN parameters with ReLU activation function to SNN with the same topology as illustrated in Figure \ref{Fig.1}, which makes it possible for SNN to obtain extremely high performance at a very low computational cost. But direct mapping will lead to severe performance degradation \cite{yang2020deterministic}. Diehl et al. \cite{diehl2015fast} propose the data-based normalization method, which scales the parameters with the maximum activation values of each layer in ANN, improving the performance of the converted SNN. Reuckauer et al.  \cite{rueckauer2017conversion} and Han et al. \cite{han2020rmp} use integrate-and-fire (IF) neurons with soft reset to make SNN achieve  performance comparable to ANN. Nonetheless, it usually takes more than 1000-4000 time steps to achieve better performance on complex datasets. And when converting ResNet \cite{he2016deep} to SNN, researchers suffer from a certain performance loss \cite{hu2018spiking,sengupta2019going,xing2019homeostasis} because the information received by the output neuron of the residual block is incomplete with the spikes on the shortcut path arriving earlier.

Bistability is a special activity form in biological neurons \cite{izhikevich2003simple}. Neurons can switch between spike and non-spike states under the action of neuromodulating substances, thus exhibiting short-term memory function  \cite{marder1996memory}. 
Inspired from the bistability characteristic, we focus on improving the performance of SNN and propose a bistable spiking neural network (BSNN), which combines phase coding and the bistability mechanism that greatly improves the performance after conversion and reduces the time delay. For high-performance spiking ResNet, we propose synchronous neurons (SN), which can help spikes in the residual block synchronously reach the output neurons from input neurons through two paths. The experimental results demonstrate they can help achieve nearly lossless conversion and state-of-the-art in MNIST, CIFAR-10, CIFAR-100, and ImageNet while significantly reduce time delay. Our contributions can be summarized as follows:
\begin{itemize}
	\item We propose a novel BSNN that combines phase coding and bistability mechanism. It effectively solves the problem of SIN and greatly reduces the performance loss and time delay of the converted SNN.
	\item We propose synchronous neurons to solve the problem that information in the spiking ResNet cannot synchronously reach the output neurons from two paths.
	\item We achieve state-of-the-art on the MNIST, CIFAR-10, CIFAR-100, and ImageNet datasets, verifying the effectiveness of the proposed method.
\end{itemize}

\section{Related Work}

Many conversion methods have been proposed in order to obtain high-performance SNN. According to the encoding method they can be divided into three kinds.

\textbf{Temporal Coding Based Conversion.} Temporal coding uses neural firing time to encode the input to spike trains and approximate activations in ANN \cite{rueckauer2018conversion}. However, since neurons in the hidden layer need to accumulate membrane potential to spike, when the activation value is equal to the maximum, neurons in deep layers are difficult to spike immediately, making this method difficult to convert deep ANNs. Zhang et al. \cite{zhang2019tdsnn} use ticking neurons to modify the method above, which transfers information layer by layer. Nevertheless, this method is less robust and difficult to be used in models with complex network structures like the residual block. 

\textbf{Rate Coding Based Conversion.} Unlike temporal coding, the rate coding-based conversion method uses the firing rates of spiking neurons to approximate the activation values in the ANN \cite{cao2015spiking}. Diehl et al.  \cite{diehl2015fast} propose data-based and model-based normalization, which use the maximum activation value of neurons in each layer to normalize the weights. When disturbed by noise, the normalization parameter may be quite large, which will cause the weight smaller and the time to spike longer. Researchers propose to use the p-th largest value for normalization operation, thereby greatly improving robustness and reducing time delay \cite{rueckauer2017conversion}.  Therefore, the conversion method based on rate coding has achieved better performance in ResNet \cite{hu2018spiking} and Inception  Networks \cite{xing2019homeostasis,sengupta2019going}.
However, the processing speed of spikes on the paths with different processing units is different. The information received by the output neuron is delayed to various degrees when spreading on these wider networks. The difference between the firing rate and the activation value in the ANN will be greater. Therefore, the performance loss and the time delay of the SNN is more significant when converting these ANNs.

\textbf{Phase Coding Based Conversion.} To overcome the large time delay of the converted SNN, researchers propose SNN with weighted spike, which assigns different weights to the spikes in different phases to pack more information in the spike \cite{kim2018deep}. Nonetheless,  when neurons do not spike in the expected phase, the spikes of neurons in hidden layers will deviate from the coding rules to a certain extent, resulting in poor performance on complex datasets and large networks. Phase coding and burst coding are combined to speed up the information transmission \cite{park2019fast}, but still needs 3000 simulation time on CIFAR-100 dataset.

\section{Proposed BSNN}

In this section, we introduce the spiking neurons and encoding methods in detail, and then analyze the reasons for the loss of conversion performance based on the process of phase coding conversion methods. The detailed information of the model to reduce conversion loss and time delay is described. And we will introduce the effect of synchronized neurons in spiking ResNet.

\subsection{Spiking Neuron and Encoding}
The most commonly used spiking neuron model is the integrate-and-fire (IF) model. The IF neuron continuously receives spikes from the presynaptic neuron and dynamically changes its membrane potential. When it exceeds the threshold, the neuron spikes and the membrane potential is traditionally reset to zero. But it will cause a lot of information loss. We follow  \cite{rueckauer2017conversion} and use the soft reset to subtract the threshold from the membrane potential:
\begin{align}
V_{i,t}^l = V_{i,t-1}^l + \sum_j w_{ij}\delta _{j,t}^{l-1},
\label{eq111}
\end{align}
\begin{align}
if \quad V_{i,t}^l \geq V_{th}, \quad \begin{cases} V_{i,t}^{l} = V_{i,t}^{l}-V_{th},\\ \delta_{i,t}^l=1. \end{cases}
\end{align}
where $V_{i,t}^l$ represents the membrane potential of neuron $i$ in layer $l$ at time $t$, $w_{ij}$ is the weight connecting the neuron $j$ and $i$, $\delta_{j,t}^{l-1}$ is the spike  of neuron $j$ in layer ($l-1$) at time $t$.

The spike trains can be encoded by real values with different encoding methods. The real value is equal to the firing rate in rate coding, which is the number of spikes in a period,  or the ratio of the spike time and total simulation time $T$ in temporal coding, which is:
\begin{align}
a_{rate}=\frac{N}{T}, \quad a_{temporal} = \frac{t_{spike}}{T},
\end{align}
where $N$ denotes the number of spikes, $t_{spike}$ is the time of the first spike. Previous work shows a considerable time delay with the use of rate and temporal coding. For example, they all need at least 1000 time steps to represent 0.001 of input. 

Therefore, we use phase coding \cite{kim2018deep} to encode activation values to spike trains. It can pack more information in one spike by assigning different weights to spikes and thresholds of each phase. Thus, phase coding is more energy efficient.  Experiments show a shorter time is taken to accurately represent the real value when phase coding is used:
\begin{align}
a_j^{l} = \frac{1}{n} \sum \limits _{k=1}^{nK}S_k \delta_{j,k}^l,\quad V_{th,t} = S_k V_{th},
\end{align}
where $a_j^l$ is the activation value of neuron $j$ in layer $l, $$K$ is the number of the phase of a period, $n=\frac{T}{K}$ is the number of the period, the phase function $S$ is represented by
\begin{align}
S_t = 2^{-(1+\mod (t, K))}.
\end{align}

\begin{figure*}[t]
	\centering 
	\includegraphics[scale=0.43]{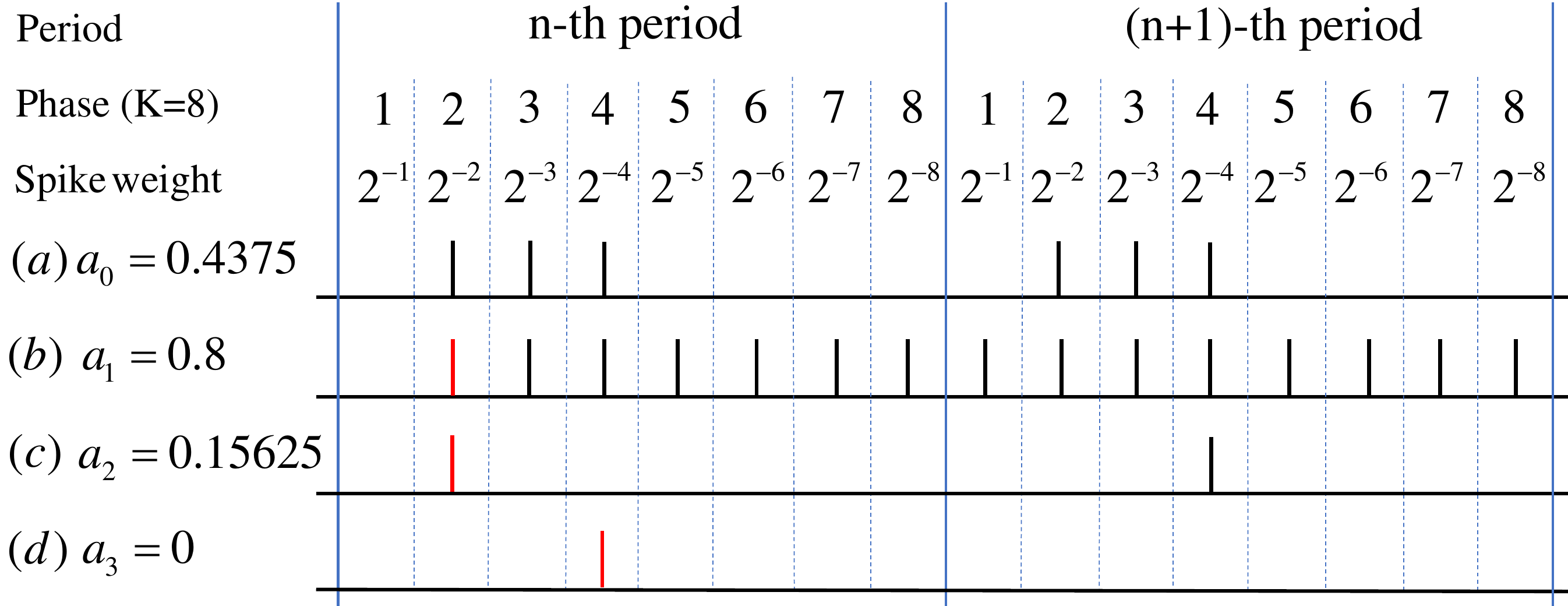}
	\caption{Examples of Phase Lag, Phase Lead, and SIN. Real value $a_0$ in (a) can be accurately represented with few spikes in phase coding. While due to the uncertainty of the postsynaptic current, phase lag problem like (b) occurs when the membrane potential in the current phase exceeds the threshold at the previous moment. When a current greater than one is suddenly received at a certain moment, phase lead problem like (c) will occur. Both of these problems will directly or indirectly lead to SIN in (d), where neurons corresponding to the activation value of zero will also spike.}
	\label{Fig.2}
\end{figure*}

\subsection{Framework of ANN-SNN Conversion}	
To make SNN work, we need to do some processing on ANN before conversion. We use  $a_i^l=\max \{0, \sum \limits_{j} w_{ij}a_j^{l-1}+b_i^l\}$ to denote the arbitrary activation value in the ANN, $w_{ij}$ and $b_i^l$ are weight and bias respectively. The maximum firing rate in SNN is one because neurons emit one spike at most at every time step. Thus, we normalize the weight and bias with the data-norm method  \cite{rueckauer2017conversion} by
\begin{align}
\label{eq1}
\hat{w}_{ij}^l = w_{ij}^l \frac{\lambda_{l-1}}{\lambda_{l}},\quad \hat{b}_{i}^l = \frac{b_i^l}{\lambda_{l}},
\end{align}
where $\hat{w}_{ij}^l $ and $\hat{b}_{i}^l $ represent the weights and biases used in SNN, $\lambda_{l}$ is the maximum activation value of the $l$-th layer. Then all activation values in ANN are at most 1.

As mentioned above, it is hard to perform max-pooling and batch normalization (BN) in SNN. We choose the spike of the neuron with the largest firing rate to output as the max-pooling operation in SNN. We follow \cite{rueckauer2017conversion} and merge the convolutional layer and the subsequent BN layer to form a new convolutional layer. An input $x$ is transformed into $BN[x]=\frac{\gamma}{\theta}(x-\mu)+\beta$, where $\mu$ and  $\theta$ are mean and variance of batch, $\beta$ and $\gamma$ are two learned parameters during training. The parameters of the new convolutional layer which can be converted, are described by
\begin{align}
\label{equ7}
\hat{w_{ij}}=\frac{\gamma_i}{\theta_i}w_{ij}, \quad \hat{b_i}=\frac{\gamma_i}{\theta_i}(b_i - \mu_i)+\beta_i.
\end{align}

\subsection{Analysis of Performance Loss}
Even though the ANN is processed, the converted SNN usually suffers performance loss.
To simplify the analysis of performance loss, we assume that $a_i^l \geq 0$, $b_i^l=0$ and the  threshold $V_{th}$ is 1. The neuron membrane potential is $V_{i,nK}^l$ at the end of the simulation. The total number of spikes of the neuron is numerically equal to the total received input minus the membrane potential at $T$:
\begin{align}
N = \sum_t^{nK}S_t\sum\limits _j w_{ij}\delta_{j,k}^{l-1} - V_{i,nK}^l.
\end{align}

Then the firing rate of neurons is approximately equal to the activation value in ANN when $T$ is long enough:
\begin{align}
r_{i,nK}^l =&\frac{N}{n}\\
=&\frac{1}{n}\ \sum_t^{nK}S_t\sum\limits _j w_{ij}\delta_{j,k}^{l-1} - V_{i,nK}^l\\
	=&\frac{1}{n}\left(\sum\limits^n\sum_j w_{ij}\sum\limits _{k=1}^K S_k\delta_{j,k}^{l-1} - V_{i,nK}^l\right)\nonumber \\
=&\sum\limits _jw_{ij}a_{j}^{l-1}-\frac{1}{n}V_{i,nK}^l.
\label{eq9}
\end{align}

Note that the postsynaptic current at each moment is as follows:
\begin{align}
I_{j,t} = \sum _jw_{ij} \delta_{j,t}^{l-1}.
\label{eq10}
\end{align}

As shown in Figure \ref{Fig.2}, once the neuron in hidden layers spikes earlier or later than the time directly encoded, which we call phase lead or phase lag, the neuron will transmit too much or too little information to the next layer. It makes some features over-activated or not activated and may cause spikes of inactivated neurons (SIN), which means neurons that cannot be activated in ANN spike in the SNN.  SNN needs a long time to accumulate spikes to reduce the impact of these destructive spikes. Thus, the features corresponding to the network firing rate can be approximately equal and proportional to the ANN features, which is the reason for the large time delay of the converted SNN. When the problem of SIN is quite severe, e.g., a large number of features that should not be activated in the ANN are activated in the SNN, it cannot be solved by long-time simulation and causes serve performance loss. Note that the above analysis is also applicable to rate-based conversion methods.

\subsection{Bistable SNN}

The immediate response of the neuron to the received current is unreliable. How should the information propagate in the spiking neurons to make the spike trains conform to the encoding rules to avoid the SIN problem caused by phase lag and phase lead? We solve the problem by proposing a bistable IF neuron (BIF) combining the IF neuron and bistability mechanism. We model the process of spiking as a piecewise function according to the fact that the bistability is shown as the periodic change of spike and non-spike states. In the spike stage, neurons spike according to the membrane potential normally while can't spike in the non-spike phase:
\begin{align}
\label{eq11}
\delta_{A,i,t}^l = \begin{cases}\mathcal H(V_{A,i,t}^l-V_{th,t}), \mod(\lfloor \frac{t}{K}\rfloor, 2)=1, \\0,\quad else. \end{cases} \nonumber\\
\delta_{B,i,t}^l = \begin{cases}\mathcal H(V_{B,i,t}^l-V_{th,t}), \mod(\lfloor \frac{t}{K}\rfloor, 2)=0, \\0,\quad else. \end{cases}
\end{align}
where $\mathcal{H}(x)$ is unit step function,  $\lfloor x \rfloor$ is the round-down operation. With periodic input, neurons do not have to respond to the input spikes all the time but accumulate spikes first and then respond and loop. Neurons respond accurately in each phase by accumulating spikes in the non-spike stage, which can effectively avoid the phase lead or lag mentioned above.

We use two BIF neurons as one unit to represent one activation value in the ANN, which is:
\begin{align}
\label{eq12}
\delta_{i,t}^l=\delta_{B,i,t}^l+\delta_{A,i,t}^l.
\end{align}

One reason for using two BIF neurons is that the BIF neuron does not spike half the simulation time. The use of two neurons with complementary spike states can make the information be transmitted to the next layer in time and maintain the continuity of information transmission. One of the neurons in two adjacent layers is in the spike state to release memory information, and the other is in the non-spike state to accumulate spikes. Note that even if the neurons in the previous layer are in the non-spike state, its silence will not interfere with the neurons in the spike state connected to the next layer. Another reason is its powerful scalability. We can convert ANNs of various topologies without carefully designing the spike stage for each layer when converting deeper and wider ANNs. If only one BIF neuron is used in each layer, when the neuron is in a spike state, it cannot play the role of accumulation as described above.

\begin{figure*}[t]
	\centering 
	\includegraphics[scale=0.5]{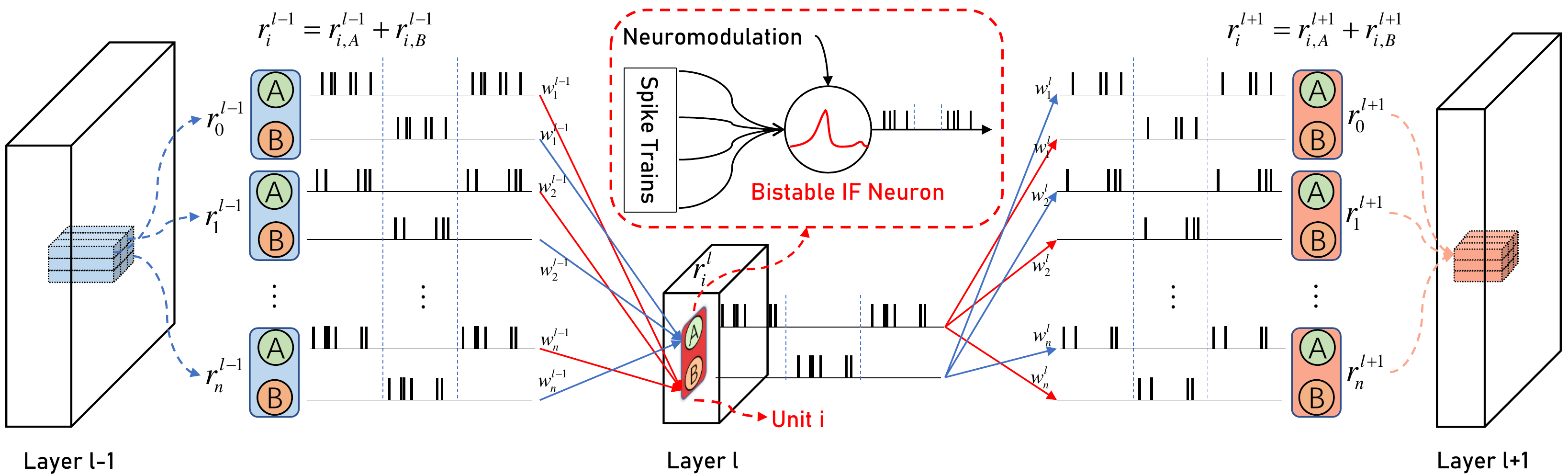}
	\caption{The Architecture of BSNN. Neuron A in the unit, which corresponds to an ANN activation value,  only has synaptic connections with neuron B in the adjacent layer and vice versa. The weights of two connections in the same unit are the same. When neuron A in the $l$-th layer is in the spike stage, neuron B in the ($l-1$)-th layer is in the non-spike stage. It will not cause interference to spikes of neuron A in layer $l$, and neuron B of the ($l+1$)-th layer can integrate the spikes until the end of the period.}
	\label{Fig:3}
\end{figure*}

As shown in Figure \ref{Fig:3}, there are two connections between the two units: neuron A of one unit is connected to neuron B of the other unit:
\begin{align}
\label{eq13}
V_{A, i,t}^l = V_{A, i,t-1}^l + \sum_j w_{ij}\delta _{B,j,t}^{l-1},\nonumber\\
V_{B, i,t}^l = V_{B, i,t-1}^l + \sum_j w_{ij}\delta _{A, j,t}^{l-1}.
\end{align}
They share the same weight. When the presynaptic neuron is in the spike phase, the postsynaptic neuron in the non-spike phase accumulates spikes to respond accurately later. In fact, the information between the two adjacent layers is periodically switched between the red connection and the blue connection with the simulation time, which also reflects that our BSNN can convert any structure of ANN.

The residual block of ResNet has two information paths, in which shortcut path connects input and output directly or through a convolution operation. The convolutional layer and the BN layer are merged to facilitate the conversion. When converting ResNet, two key problems need to be addressed:
\begin{itemize}
	\item \textbf{The information of the two paths cannot be scaled synchronously.} The information of two paths received by output neurons of the residual block is not proportional to the activation values. Because it is impossible to normalize the shortcut path which has no convolutional layer.
	\item \textbf{The information of the two paths cannot reach the output neuron synchronously.}  
	The shortcut path is one less ReLU operation,  which corresponds to two BIF neurons in the SNN,  than the convolution path. Since neurons need time to accumulate membrane potential to spike, the information of the shortcut path reaches the output neuron faster.
\end{itemize}

\subsection{Synchronous Neurons for Spiking ResNet}
\begin{figure}[bp]
	\centering 
	\includegraphics[scale=0.40]{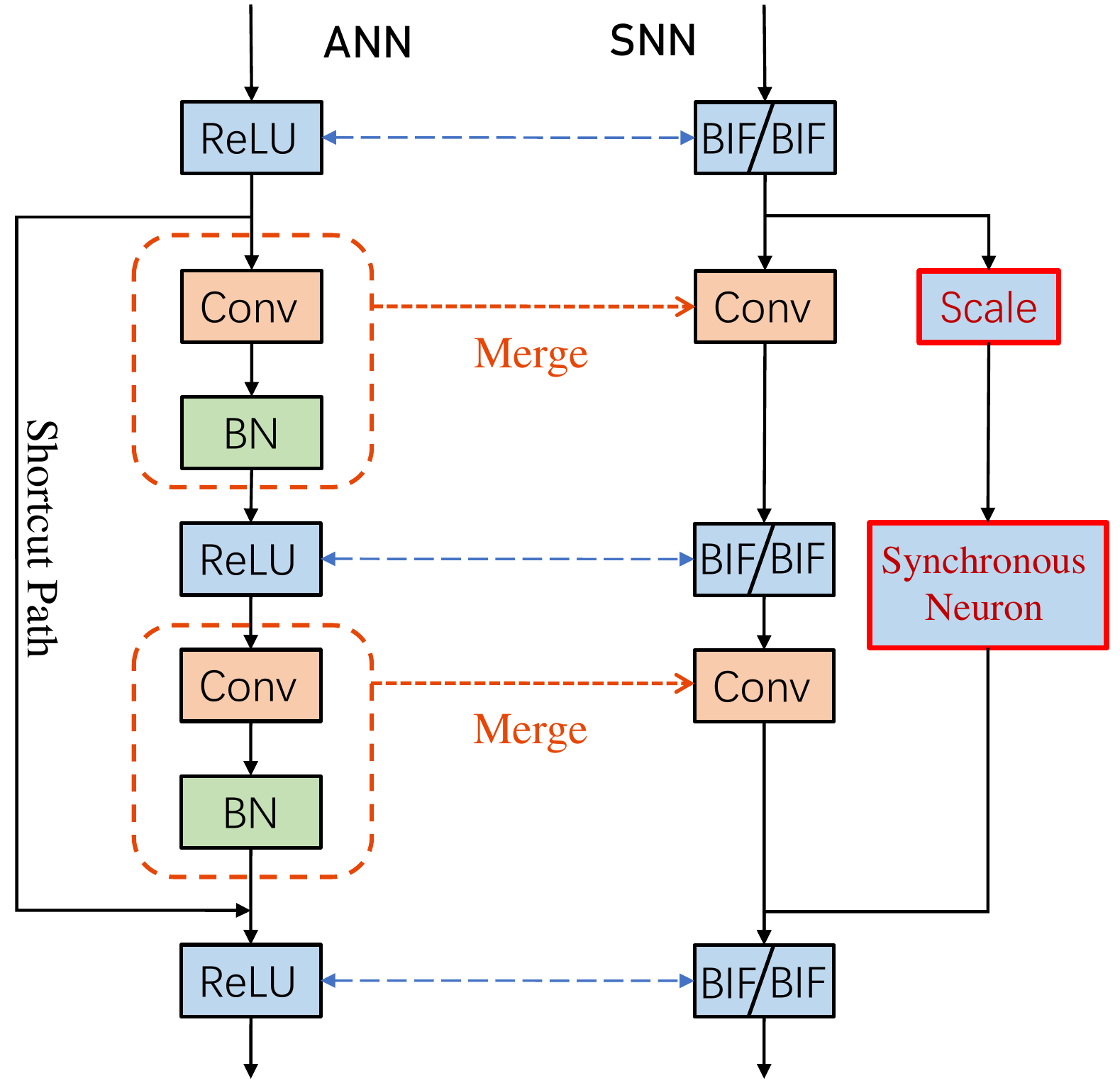}
	\caption{Synchronous Neurons for Spiking ResNet}
	\label{Fig:4}
\end{figure}

For the first problem, we determine the scale parameters according to the maximum activation value of the input and output so that the sum of the information of the two paths received by the output is proportional to the activation value:
\begin{align}
scale = \frac{\lambda_{in}}{\lambda_{out}}.
\end{align}

To solve the second problem, we add synchronous neurons, which are two BIF neurons, in the shortcut path. It is equivalent to adding a ReLU function to the head of the shortcut path in ANN. Figure \ref{Fig:4} shows the conversion process of the residual block. The information reaches the output of the residual block through the synchronous neurons. Since the input of the shortcut path is all non-negative, the transmission in ANN will not have any impact. In SNN, due to the existence of synchronous neurons, the output of the shortcut path and the convolutional path will reach the output neuron at the same time, thereby eliminating the phase lead and lag and SIN problems in spiking ResNet.

The entire conversion process summarized in Algorithm \ref{alg:algorithm} where the SNNs transmit information with BIF neurons. 

\begin{algorithm}[bp]
	\caption{ANN-SNN Conversion with BIF Neurons}
	\label{alg:algorithm}
	\textbf{Input}: Training and test set, simulation time $T$, trained ANN\\
	\textbf{Output}: Performance of the SNN
	\begin{algorithmic}[1] 
		\STATE Let $V_{th}=1, \lambda_l=0$ for $l=1, \cdots, L$ to save the maximum activation value of each ANN layer.
		\STATE Merge the convolutional layer and BN layer according to equation (\ref{equ7}).
		\FOR {$l=1$ to $L$}
		\STATE $a^l \leftarrow$ layer-wise activation value
		\STATE $\lambda_l=\max\{a_i^l\}$
		\ENDFOR
		\FOR {$l=1$ to $L$}
		\STATE $\hat{w}_{ij}^l = w_{ij}^l \frac{\lambda_{l-1}}{\lambda_{l}}, \quad \hat{b}_{i}^l = \frac{b_i^l}{\lambda_{l}}$
		\ENDFOR
		\STATE Map the processed parameters to the SNN.
		\FOR {$s=1$ to $\#$ of test set}
		\FOR {$t=1$ to $L$}
		\STATE do inference according to equation (\ref{eq11})(\ref{eq12})(\ref{eq13})
		\ENDFOR
		\ENDFOR
		\STATE \textbf{return} performance of the SNN
	\end{algorithmic}
\end{algorithm}

\section{Experiment}

In this section, various experiments are conducted to evaluate the performance of our proposed conversion algorithm. We also test the effect of the synchronous neurons and compare our BSNN with various advanced conversion algorithms.

\subsection{Dataset}
The MNIST \cite{lecun1998gradient}, CIFAR-10, CiFAR-100 \cite{krizhevsky2009learning}, and ImageNet \cite{deng2009imagenet} datasets are used to test the performance of our proposed BSNN.

The MNIST dataset is the most commonly used dataset and benchmark for classification tasks. It contains 60,000 handwritten digital images from 0 to 9, 50,000 images for the training set, and 10,000 images for the test set. Each image contains 28x28 pixels, which are represented in the form of 8-bit gray values. Note that we do not perform any preprocessing on the MNIST dataset.

The CIFAR-10 dataset is the color image dataset closer to universal objects and a benchmark test set of the CNN architecture. It contains 60,000 images of 10 classes. 50,000 images for the training sets, and 10,000 images for the test sets. It is a 3-channel color RGB image, whose size of each image is 32x32. Unlike MNIST, we normalize the dataset to make the CIFAR-10 obey a standard normal distribution.

The CIFAR-100 dataset has the same image format as CIFAR-10. We also perform the same normalization operation on it, with different normalization parameters. The difference with CIFAR-10 is that CIFAR-100 contains 100 categories instead of 10. Each category contains 500 training images and 100 test images.

ImageNet is currently the world's largest image recognition large-scale labeled image database organized according to the wordnet structure, and it is also the most challenging classification dataset for SNN. Among them, the training set is 1281167 pictures, and the verification set is 50,000 pictures, including 1000 different categories and 3-channel natural images. The normalization process is also performed to obtain a sufficiently high classification performance.

\begin{figure*}[bp]
	\centering
	\subfigure[Rate-based conversion]{
		\label{Fig(a)} 
		\includegraphics[scale=0.44]{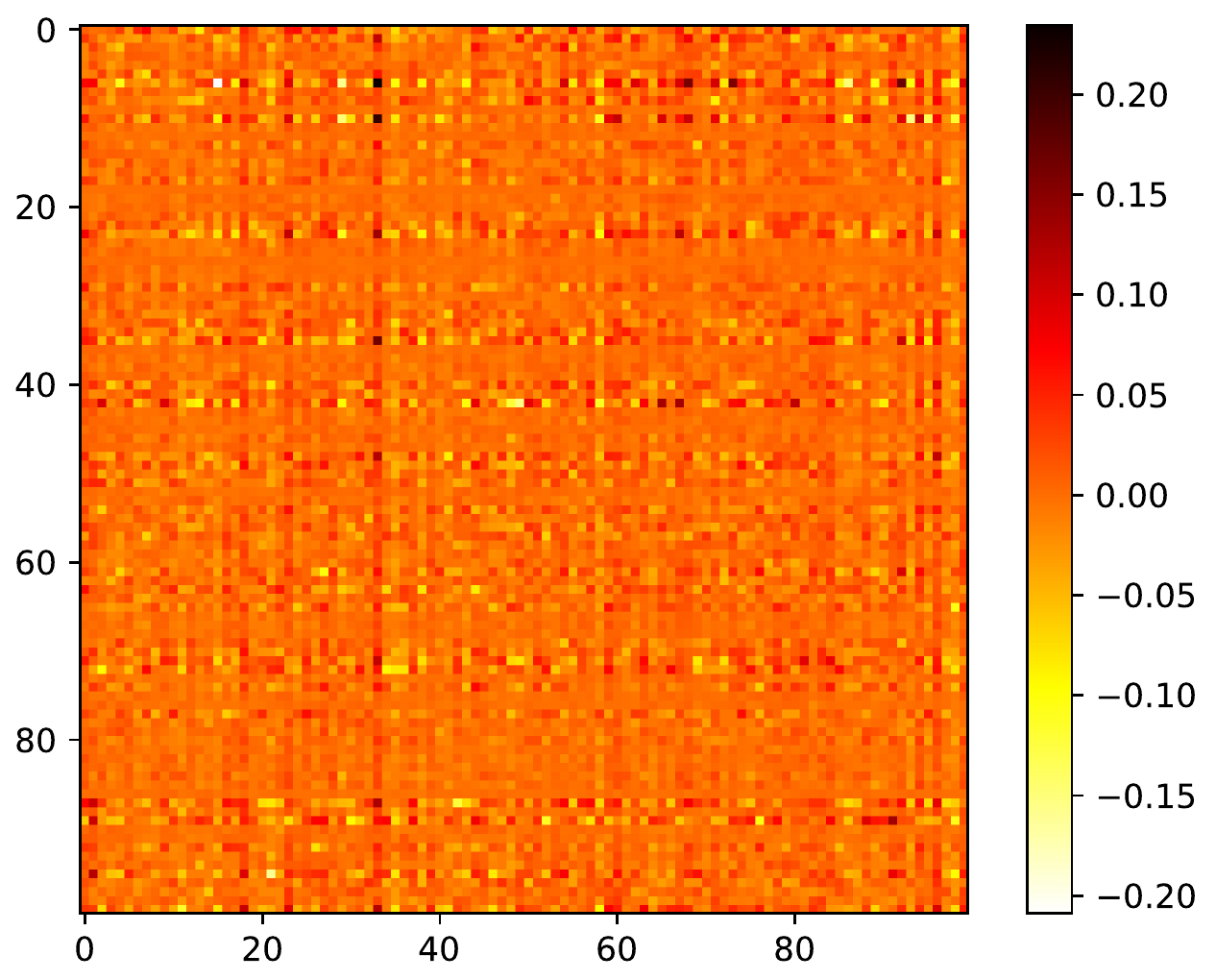}}
	\subfigure[Phase-based conversion]{
		\label{Fig(b)} 
		\includegraphics[scale=0.44]{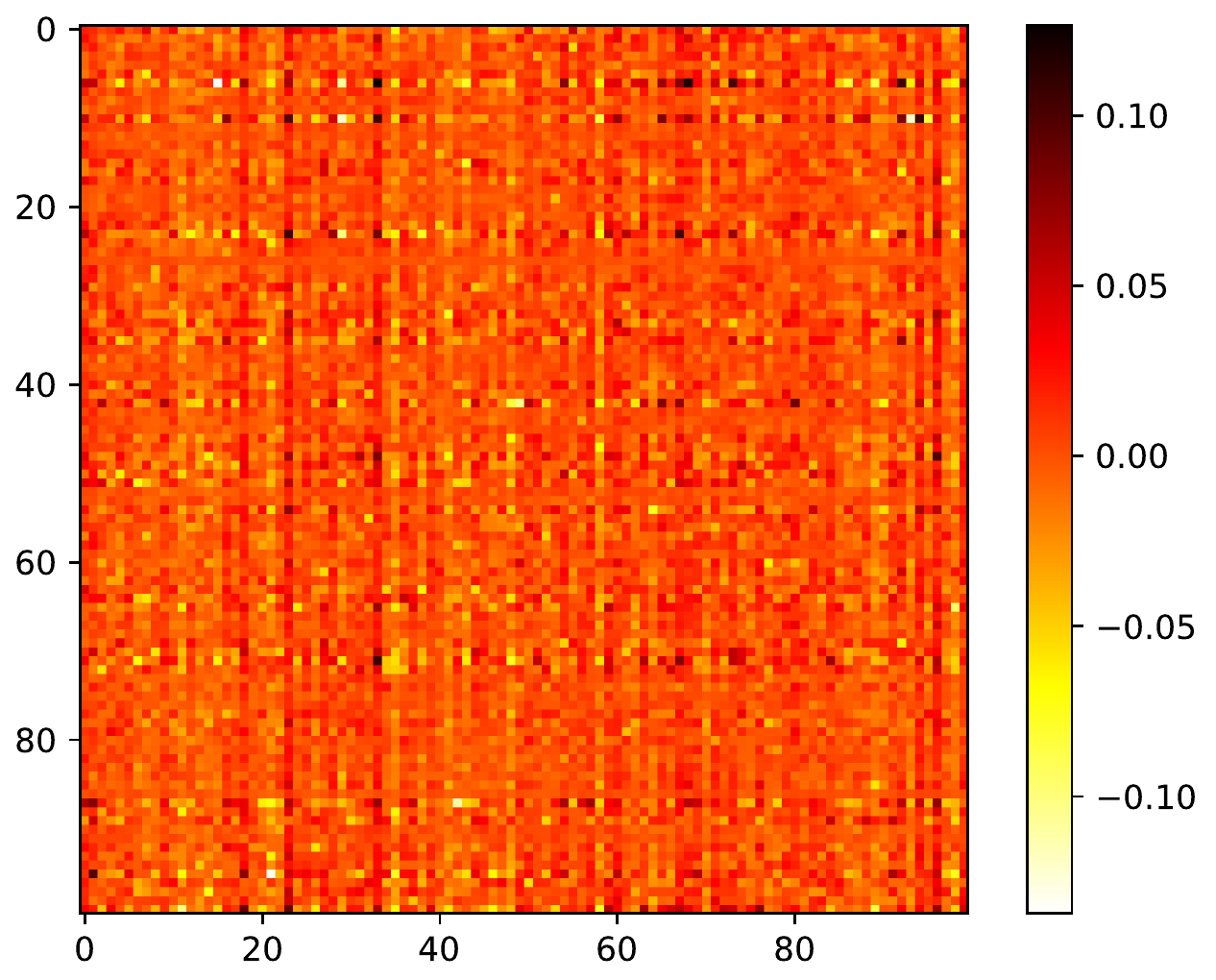}}
	\subfigure[BSNN]{
		\label{Fig(c)} 
		\includegraphics[scale=0.44]{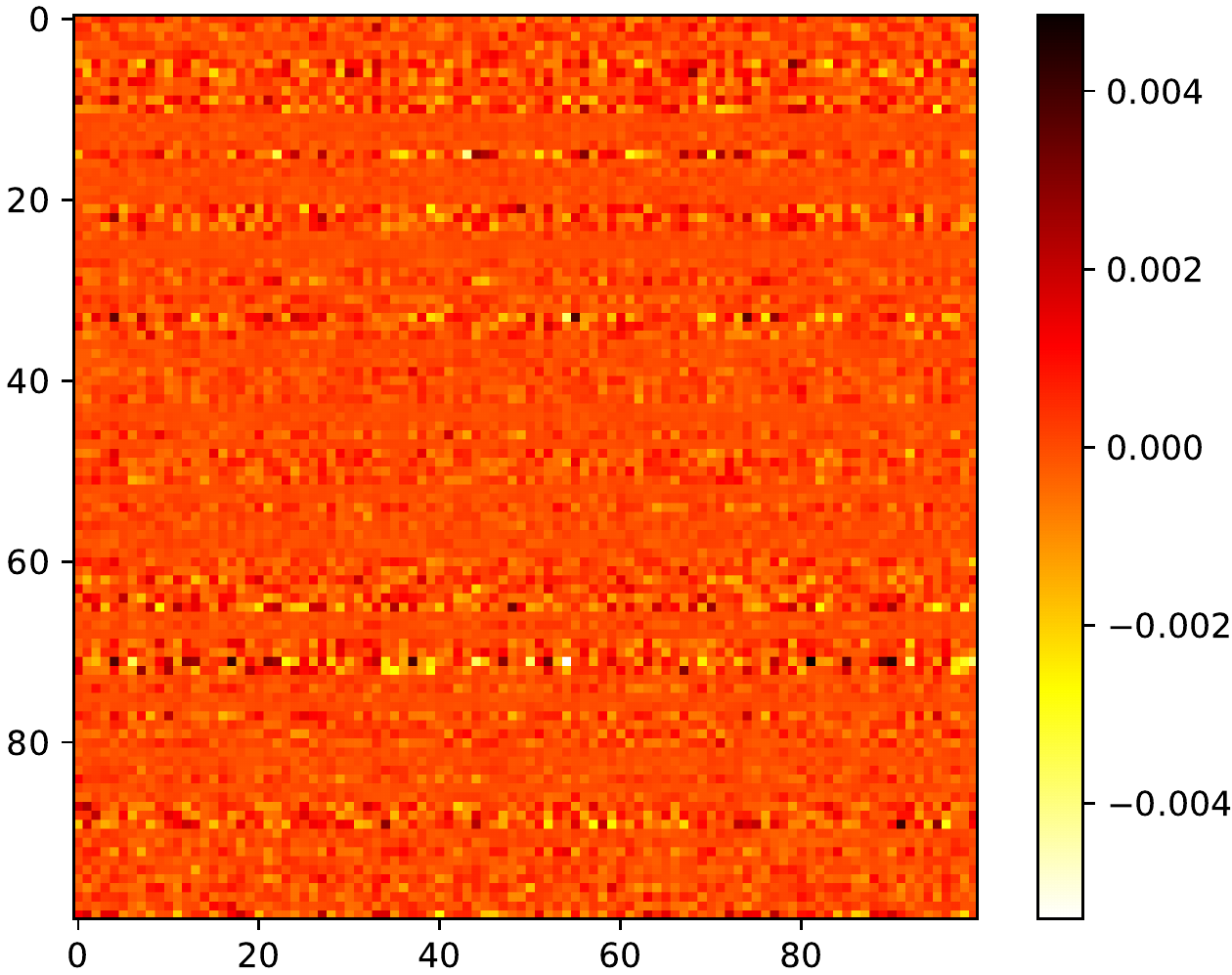}}
	\caption{Output difference between ANN and converted SNN on CIFAR-100. We compare the output rate of 100 samples of VGG16 with the output firing rate of three conversion methods to verify that the reason for BSNN to improve network performance lies in better approximation.}
	\label{Fig:5} 
\end{figure*}

\subsection{Experimental Setup}
Our experiments are implemented on the Pytroch framework and NVIDIA A100. We convert CNN with 12c5-2s-64c5-2s-10 architecture \cite{kim2018deep} on MNIST. 12c5 means a convolutional layer with 12 output channels and kernel size of 5 and 2s refers to non-overlapping pooling layer with kernel size of 2. We use VGG16, ResNet18, ResNet20 architecture on CIFAR-10 and CIFAR-100, while ResNet18 and ResNet34 are used for experiments on ImageNet. Their structures are the same as that of Pytorch's built-in model. We train the ANN for 100 or 300 epochs by using the stochastic gradient descent algorithm. The initial learning rate is 0.01, and the learning rate is scaled by 0.1 at the training epoch of [180, 240, 270]. Other parameters are listed in Table \ref{canshu}. We perform data augmentation on the input except for the MNIST dataset and use real-value input in SNNs for better performance. When comparing various conversion methods, except for the encoding and the way of information transmission, the other settings are the same. 

\begin{table}[h]
	\centering
	\begin{tabular}{lrrrr}
		\toprule
		Parameters & MNIST & CIFAR-10 & CIFAR-100 & ImageNet\\
		\midrule
		training epoch & 100 & 300 & 300 &300\\
		total time step & 80 & 400 & 800& 1000\\
		batch size & 10000 & 100 & 100 & 50\\
		p-max & 0.999 & 1.0 & 1.0 & 1.0\\
		threshold & 1.0 & 1.0 & 1.0 & 1.0\\
		\bottomrule
	\end{tabular}
	\caption{Parameters of BSNN}
	\label{canshu}
\end{table}

\subsection{Performance and Comparsion with other Methods}

To obtain a high-performance SNN, the firing rate of the converted SNN should be similar or equal to the activation value of ANN, which is consistent with the conversion principle.
We check the output difference of 100 samples of CIFAR-100 between the firing rate of converted SNN and the corresponding activation value of the ANN with architecture of VGG16. Ideally, due to the weight normalization, the output of the ANN is proportional to the firing frequency of the SNN output, and the multiple is the maximum value of the ANN output layer. We multiply the output of the SNN with the multiple for comparison. As we can see from Figure \ref{Fig:5}, the difference between the output of the selected 100 samples and the output of the ANN is mostly near 0. However, although the rate-bsed conversion method is widely used, it can be seen from the output of the network that the performance loss is that SNN cannot approach the activation value of ANN very well. The method based on phase encoding reduces the difference between the outputs by increasing the amount of information contained in the spikes, however, the problem of inaccurate approximation is still not solved. As can be seen in Figure \ref{Fig(c)}, the output of BSNN is at most 0.005 different from the corresponding activation value of ANN. This indicates that the improvement of performance with BSNN comes from the accurate approximation to ANN activation values.

\begin{table*}[t]
	\centering
	\begin{tabular}{llllrrrr}
		\toprule
		Dataset & Method & Network & Encoding&  ANN (\%) & SNN (\%) & Loss (\%) & Time Steps\\
		\midrule
		& p-Norm \cite{rueckauer2017conversion}    & CNN  & Rate&  99.44  & 99.44  & 0.00     & -   \\
		MNIST& Weighted Spikes \cite{kim2018deep}    & CNN& Phase  &  99.20  & 99.20   & 0.00      & 16  \\
		& \textbf{BSNN}     & \textbf{CNN}  & \textbf{Phase}&  \textbf{99.30}   &\textbf{99.31}   & \textbf{-0.01}  & \textbf{35}  \\
		\midrule
		& p-Norm \cite{rueckauer2017conversion}   & VGG16& Rate &91.91  & 91.85 & 0.06 & 35\\
		& Spike-Norm \cite{sengupta2019going} & VGG16& Rate & 91.70 & 91.55&0.15 & -\\
		& Hybrid Training \cite{rathi2020enabling} & VGG16& Rate & 92.81 & 91.13 & 1.68 & 100\\
		& RMP-SNN \cite{han2020rmp} & VGG16& Rate & 93.63 & 93.63 & 0.00 & 1536\\
		& TSC \cite{han2020deep} & VGG16 & Temporal& 93.63 & 93.63 & 0.00 & 2048\\
		& CQ Trained \cite{yan2021near} & VGG16 & Rate & 92.56 & 92.48 & 0.08 & 600\\
		CIFAR-10 & \textbf{BSNN} & \textbf{VGG16} & \textbf{Phase}& \textbf{94.11} & \textbf{94.12} & \textbf{-0.01} & \textbf{166}\\
		& Weighted Spikes \cite{kim2018deep} & ResNet20 & Phase& 91.40 & 91.40 & 0.00 & -\\
		& Hybrid Training \cite{rathi2020enabling} & ResNet20 & Rate& 93.15 & 92.22 & 0.93 & 250\\
		& RMP-SNN \cite{han2020rmp} & ResNet20 & Rate& 91.47 & 91.36 & 0.11 & - \\
		& TSC \cite{han2020deep} & ResNet20 & Temporal& 91.47 & 91.42 & 0.05 & 1536\\
		& \textbf{BSNN} & \textbf{ResNet20} & \textbf{Phase}& \textbf{95.02} & \textbf{95.16} & \textbf{-0.14} & \textbf{206}\\	
		\midrule
		& Hybrid Training \cite{rathi2020enabling} & VGG11& Rate & 71.21 & 67.87 & 3.34 & 125\\
		& RMP-SNN \cite{han2020rmp} & VGG16 & Rate& 71.22 & 70.93 & 0.29 & 2048 \\
		& TSC \cite{han2020deep} & VGG16 & Temporal& 71.22 & 70.97 & 0.25 & 1024\\
		& CQ Trained \cite{yan2021near} & VGG & Rate & 71.84 & 71.84 & 0.00 & 300\\
		& \textbf{BSNN} & \textbf{VGG16} & \textbf{Phase}& \textbf{73.26} & \textbf{73.41} & \textbf{-0.15} & \textbf{242}\\
		CIFAR-100& Spiking ResNet \cite{hu2018spiking}& ResNet44 & Rate& 70.18 & 68.56 & 1.62 & -  \\
		& Weighted Spikes \cite{kim2018deep} & ResNet32 & Phase& 66.10 & 66.20 & -0.10 &- \\
		& RMP-SNN \cite{han2020rmp} & ResNet20 & Rate& 68.72 & 67.82 & 0.90 & 2048 \\
		& TSC \cite{han2020deep} & ResNet & Temporal& 68.72 & 68.18 & 0.54 & 2048\\
		& \textbf{BSNN} & \textbf{ResNet20}& \textbf{Phase} & \textbf{77.97} & \textbf{78.12} & \textbf{-0.15} & \textbf{265}\\
		\midrule
		& Spike-Norm \cite{sengupta2019going} & ResNet20 & Rate& 70.52 & 69.39 & 1.13 & -\\
		& \textbf{BSNN} & \textbf{ResNet18} & \textbf{Phase}& \textbf{69.65} & \textbf{69.65} & \textbf{0.00} & \textbf{200}\\
		ImageNet & Hybrid Training \cite{rathi2020enabling} & ResNet34 & Rate& 70.20 & 61.48 & 8.72 & 250 \\
		& RMP-SNN \cite{han2020rmp} & ResNet34 & Rate& 70.64 & 69.89 & 0.75 & 4096\\
		& \textbf{BSNN} & \textbf{ResNet34} & \textbf{Phase}& \textbf{73.27} & \textbf{72.64} & \textbf{0.63} & \textbf{989}\\
		\bottomrule
	\end{tabular}
	\caption{Top-1 classification accuracy on MNIST, CIFAR-10, CIFAR-100 and ImageNet for our converted SNNs, compared to the original ANNs, and compared to other conversion methods}
	\label{tab:2}
\end{table*}

Then we compare the performance of our model and other conversion methods on MNIST, CIFAR-10, CIFAR-100, and ImageNet, as shown in Table \ref{tab:2}. The time step is the simulation time required to achieve the best performance. We choose rate-based methods including p-Norm \cite{rueckauer2017conversion}, Spike-Norm \cite{sengupta2019going}, RMP-SNN \cite{han2020rmp}, etc., phase-based Weighted Spikes \cite{kim2018deep} method, temporal coding-based TSC \cite{han2020deep} method, and other advanced methods such as CQ trained \cite{yan2021near}, Hybrid training \cite{rathi2020enabling}, etc. for comparison. Here we do not compare the BSNN with algorithms based on biological rules and backpropagation. Because the former focuses on the biological interpretability of the network, while the latter focuses on exploring the temporal and spatial representation of features. The training cost of both is particularly high because of the information processing method similar to RNN in the training process. It is difficult to apply them to complex networks such as VGG16 and ResNet34, Thus, their performance significantly lags behind advanced conversion-based methods.

We first focus on the performance loss of the conversion method. The phase-based method is usually better than other methods because it combines the advantages of rate coding and temporal coding. The time information expressed in phase and the rate information expressed in period improve the information expressing ability of the spike. Based on this, our BSNN improves the information propagation of SNN based on BIF neurons and reduces the phase lead and lag problems in the Weighted Spike method, thus minimizing the performance loss. We achieved 99.31\% performance on MNIST, 94.12\% (VGG16), and 95.02\%  (ResNet20) performance on CIFAR-10, 73.41\% (VGG16) and 78.12\% (ResNet20) performance on CIFAR-100, and 69.65\% (ResNet18) performance on ImageNet, which are better than other conversion method. To continue testing the ability of our method to convert deep networks, we conduct experiments on ResNet34. The results show that BSNN only needs less than 1000 time steps to achieve the performance of 72.64\% with only 0.63\% performance loss. As far as we know, this is also the highest performance that SNN can achieve.

In addition to the excellence in accuracy, our model has also achieved outstanding performance in time steps. The conversion method based on rate and timing naturally takes a long time to accurately represent the information and therefore requires a longer time step. The Hybrid Training method sacrifices part of the performance in exchange for shorter simulation time. We analyze above that the reason why the conversion method requires a long simulation time is that SNN needs enough spikes to compensate for the destruction of the proportional relationship caused by spikes of inactivated neurons. BSNN uses the bistable mechanism to accumulate and release spikes, thus the SIN problem is significantly improved. As shown in the Table \ref{tab:2}, on complex data sets such as CIFAR-10 and ImageNet, BSNN only needs a time step of 1/4 to 1/10 to achieve the performance of other advanced algorithms. hence, BSNN can save at least 25\% of calculation loss and energy consumption to a certain extent, which plays an important role in the development and application of SNN.

\subsection{Effect of Synchronous Neuron}

\begin{figure}[h]
	\centering
	\includegraphics[scale=0.38]{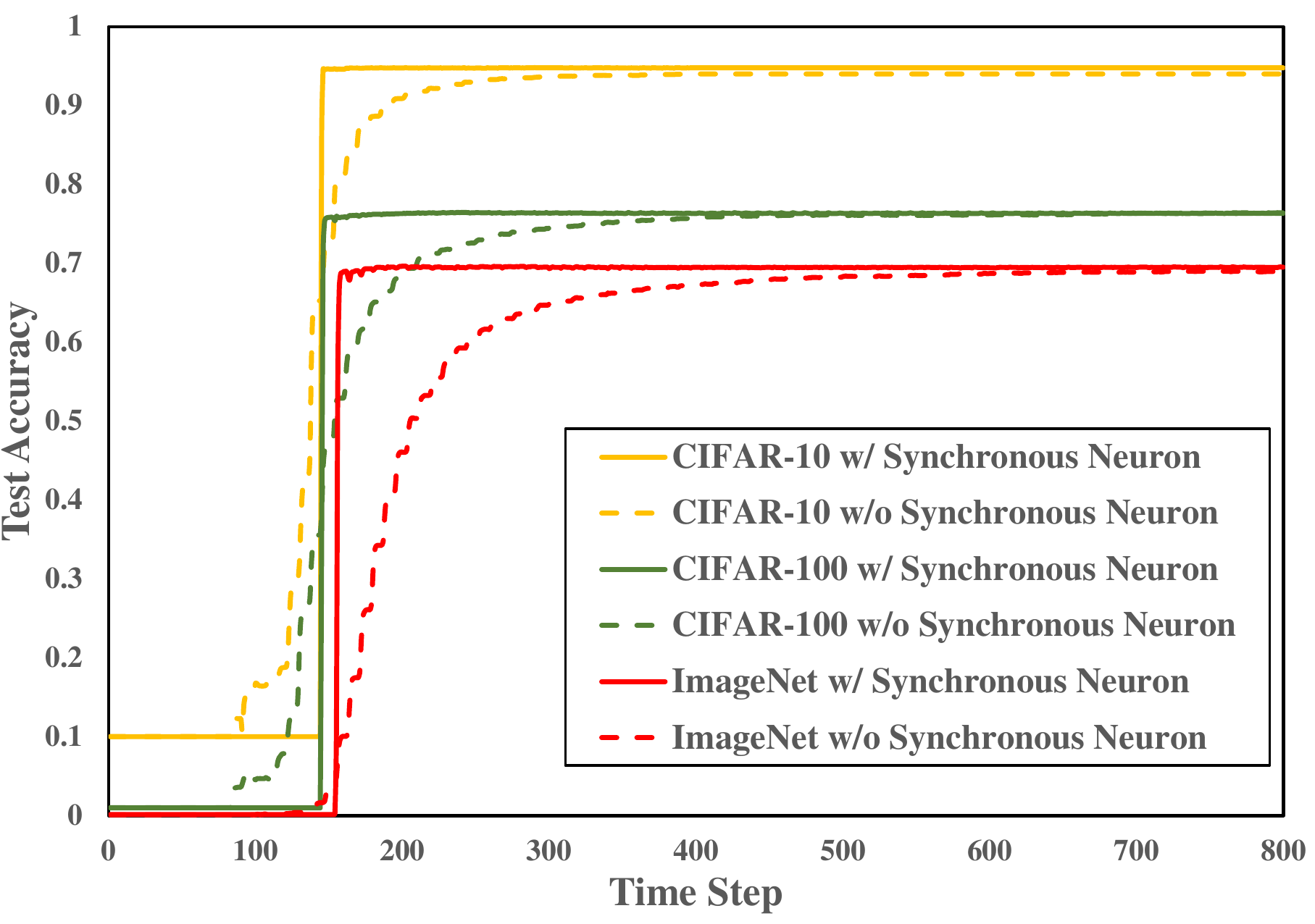}
	\caption{The Performance of  Spiking ResNet18 on three datasets.}
	\label{Fig:6} 
\end{figure}

In order to verify the effectiveness of the proposed synchronous neuron in converting ResNet, we convert ResNet18 on multiple datasets. As shown in Figure \ref{Fig:6} \footnote{We only run 400 time steps on the CIFAR-10 dataset. For the convenience of drawing, we use the result of step 400 for the next 400 time steps.}, since neurons are not always in the spike state but switch between spike and non-spike states, BSNN doesn't work in the early simulation but completes the high-precision conversion with a small time delay. The detailed results are listed in Table \ref{tab:1}. The loss means the accuracy difference ($acc_{ANN}-acc_{SNN}$) between the source ANN and the converted SNN. The experimental results show that the performance of the spiking ResNet using synchronous neurons exceeds the SNNs without synchronous neurons on CIFAR-10, CIFAR-100, and ImageNet datasets. It achieves the same performance as the ANN with 200-800 time-steps reduction. The use of synchronous neurons on ResNet conversion can ensure that the information of two paths reaches the output neuron of the residual block synchronously, which significantly improves the conversion accuracy and reduces the time delay.

\begin{table}[t]
	\centering
	\begin{tabular}{llrrr}
		\toprule
		& Dataset  &  SNN (\%)& Loss (\%) & Time\\
		\midrule
		& CIFAR-10 & 94.04 & 0.74 &395\\
		w/out SN & CIFAR-100 &  76.37 & 0.03 & 741\\
		& ImageNet & 69.32 & 7.08 & 996\\
		
		\midrule
		& CIFAR-10 &  94.83 & -0.05 &218\\
		w/ SN & CIFAR-100 &  76.48 & -0.08 & 237\\
		& ImageNet &  69.64 & 0.00 & 200\\
		\bottomrule
	\end{tabular}
	\caption{The Results of Adding Synchronous Neurons on ResNet18. }
	\label{tab:1}
\end{table}

Note that previous work like Spike-Norm \cite{sengupta2019going} uses average pooling and dropout instead of max-pooling and BN, limiting the performance of the converted SNN to a certain extent. The results show that our work can be adapted to various types of ANNs, and achieve almost lossless conversion with less time delay. 
Experimental results on complex datasets like CIFAR-100 and deep networks like ResNet34 show that BSNN can solve the difficulty in approximating features in deep layers to ANN by cooperating two BIF neurons of each unit to accumulate and emit spikes periodically. It means that we can achieve the same effect as current deep learning with a more biologically plausible network structure, less computational cost and energy consumption.

\section{Conclusion}

In this paper, we analyze the reasons for the performance loss and large time delay in the conversion method. Our analysis reveals that the immediate response of neurons to the received current is unreliable in converted SNNs. It can bring the problem of SIN, which makes the firing rate in the deep layer cannot approximate the activation values in ANNs. Based on these analysis and observation, we propose a novel Bistable SNN which combines phase coding and the bistability mechanism, and design synchronous neurons to improve energy-efficiency, performance, and inference speed.  Our experiments demonstrate that the BSNNs could significantly reduce performance loss and time delay. The efficiency and efficacy of our proposed BSNN could thus be of great importance for fast and energy-efficiency spike-based neuromorphic computing.

\section*{Acknowledgments}
This work is supported by the National Key Research
and Development Program (2020AAA0107800), the Strategic Priority Research Program of the Chinese Academy of Sciences (Grant No. XDB32070100), the Beijing Municipal Commission of Science and Technology (Grant No. Z181100001518006), and the Beijing Academy of Artificial Intelligence (BAAI).


%

\ifCLASSOPTIONcaptionsoff
  \newpage
\fi



\bibliographystyle{IEEEtran}
%
%
\bibliography{BSNN}

%

\begin{IEEEbiography}[{\includegraphics[width=1in,height=1.25in,clip,keepaspectratio]{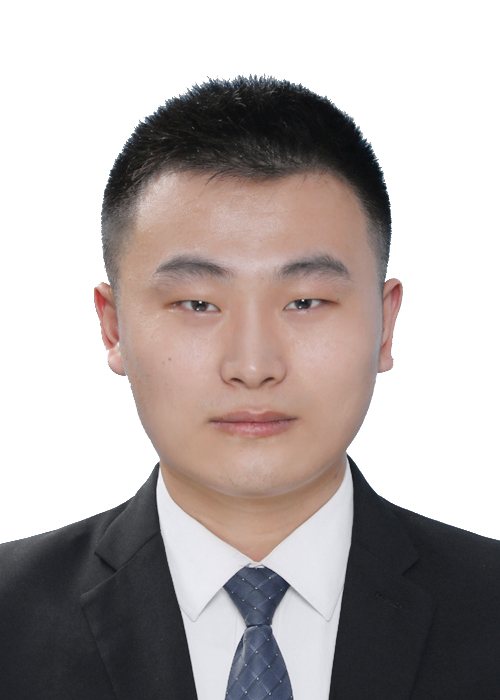}}]{Yang Li}
received the B.Eng. degree in automation from Harbin Engineering University, Harbin, China, in 2019. He is currently pursuing the master’s degree with the Institute of Automation, Chinese Academic of Sciences, Bejing, China. His current research interests include learning algorithms in spiking neural networks and cognitive computations.
\end{IEEEbiography}

\begin{IEEEbiography}[{\includegraphics[width=1in,height=1.25in,clip,keepaspectratio]{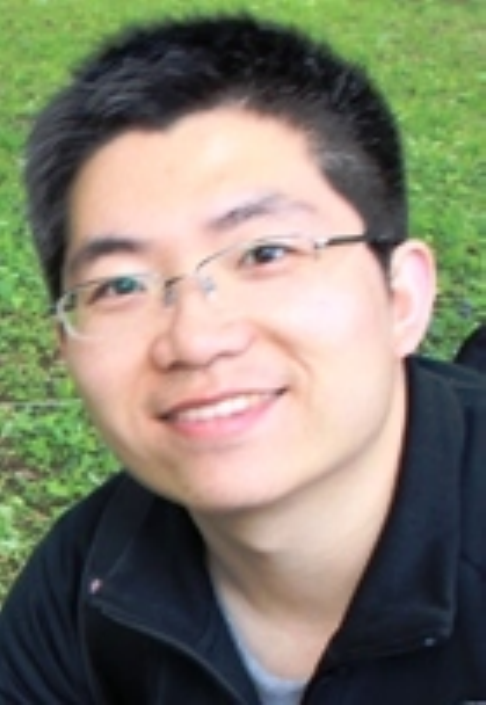}}]{Yi Zeng}
obtained his Bachelor degree in 2004 and Ph.D degree in 2010 from Beijing University of Technology, China. He is currently a Professor and Deputy Director at Research Center for Brain-inspired Intelligence, Institute of Automation, Chinese Academy of Sciences (CASIA), China. He is also with the National Laboratory of Pattern Recognition, CASIA, and University of Chinese Academy of Sciences, China. He is a Principal Investigator at Center for Excellence of Brain Science and Intelligence Technology, Chinese Academy of Sciences,
China. His research interests include cognitive brain computational modeling, brain-inspired neural networks, brain-inspired robotics, etc.
\end{IEEEbiography}


\begin{IEEEbiography}[{\includegraphics[width=1in,height=1.25in,clip,keepaspectratio]{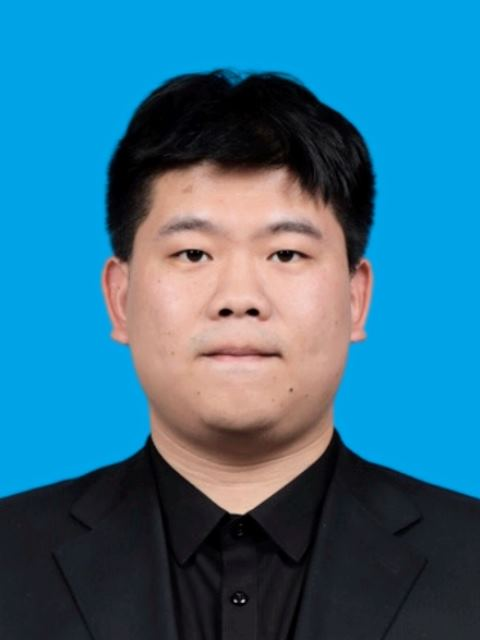}}]{Dongcheng Zhao}
received the B.Eng. degree in Information and Computational Science from XiDian University, Xi’an, Shaanxi, China, in 2016. He is currently pursuing the PH.D. degree with the Institute of Automation, Chinese Academic of Sciences, Bejing, China. His current research interests include learning algorithms in spiking neural networks, thalamus-cortex interaction, visual object tracking, etc.
\end{IEEEbiography}




\end{document}